\documentclass{nature_mi/sn-jnl}%

\usepackage{graphicx}%
\usepackage{multirow}%
\usepackage{amsmath,amssymb,amsfonts}%
\usepackage{amsthm}%
\usepackage{mathrsfs}%
\usepackage[title]{appendix}%
\usepackage{xcolor}%
\usepackage{textcomp}%
\usepackage{manyfoot}%
\usepackage{booktabs}%
\usepackage{listings}%
\usepackage{xspace}
\usepackage[capitalise]{cleveref}
\usepackage{hyperref} %
\usepackage[inkscapeformat=png]{svg}
\usepackage{multirow}
\usepackage{colortbl}
\usepackage{anyfontsize}
\usepackage{tablefootnote}
\usepackage{algorithm}%
\usepackage{algorithmicx}%
\usepackage{algpseudocode}%

\usepackage{xr} %
\DeclareUnicodeCharacter{2009}{\,}
\makeatletter
\DeclareRobustCommand\onedot{\futurelet\@let@token\@onedot}
\def\@onedot{\ifx\@let@token.\else.\null\fi\xspace}

\newcommand*{\addFileDependency}[1]{%
\typeout{(#1)}%
\@addtofilelist{#1}
\IfFileExists{#1}{}{\typeout{No file #1.}}
}

\makeatother

\newcommand{\zbf}{\mathbf{z}}
\newcommand{\xbf}{\mathbf{x}}
\newcommand{\bbf}{\mathbf{b}}

\newcommand{\Wbf}{\mathbf{W}}
\newcommand{\Cov}{\mathrm{Cov}}
\newcommand{\diag}{\mathrm{diag}}
\definecolor{uclablue}{rgb}{0.15, 0.45, 0.68}
\definecolor{aliceblue}{RGB}{178, 217, 245}
\definecolor{babyblue}{RGB}{217, 239, 251}
\definecolor{babypink}{RGB}{251, 231, 230}
\newcommand{\CC}{\cellcolor{aliceblue}}
\newcommand{\CP}{\cellcolor{babypink}}

\begin{document}

\title[Functional groups are all you need]{Functional Groups are All you Need for Chemically Interpretable Molecular Property Prediction}

\author[1,2,3]{\fnm{Roshan} \sur{Balaji}}\email{roshan@smail.iitm.ac.in}

\author[1]{\fnm{Joe} \sur{Bobby}}\email{joebobby72@gmail.com}

\author*[1,2,3,4]{\fnm{Nirav} \sur{Pravinbhai Bhatt}}\email{niravbhatt@iitm.ac.in}

\affil[1]{\orgdiv{BioSystems Engineering and Control Lab}, \orgname{Indian Institute of Technology Madras}, \orgaddress{\city{Chennai}, \state{Tamil Nadu}, \country{India}}}

\affil[2]{\orgdiv{The Centre for Integrative Biology and Systems medicinE (IBSE)}, \orgname{Indian Institute of Technology Madras}, \orgaddress{\city{Chennai}, \state{Tamil Nadu}, \country{India}}}

\affil[3]{\orgdiv{Wadhwani School of Data Science and AI}, \orgname{Indian Institute of Technology Madras}, \orgaddress{\city{Chennai}, \state{Tamil Nadu}, \country{India}}}

\affil[4]{\orgname{Indian Institute of Technology Madras Zanzibar}, \orgaddress{\state{Zanzibar}, \country{Republic of Tanzania}}}

\abstract{Molecular property prediction using deep learning (DL) models has accelerated drug and materials discovery, but the resulting DL models often lack interpretability, hindering their adoption by chemists. This work proposes developing molecule representations using the concept of Functional Groups (FG) in chemistry. We introduce the Functional Group Representation (FGR) framework, a novel approach to encoding molecules based on their fundamental chemical substructures. Our method integrates two types of functional groups: those curated from established chemical knowledge (FG), and those mined from a large molecular corpus using sequential pattern mining (MFG). The resulting FGR framework encodes molecules into a lower-dimensional latent space by leveraging pre-training on a large dataset of unlabeled molecules. 
Furthermore, the proposed framework allows the inclusion of 2D structure-based descriptors of molecules. 
We demonstrate that the FGR framework achieves state-of-the-art performance on a diverse range of 33 benchmark datasets spanning physical chemistry, biophysics, quantum mechanics,  biological activity, and pharmacokinetics while enabling chemical interpretability. Crucially, the model's representations are intrinsically aligned with established chemical principles, allowing chemists to directly link predicted properties to specific functional groups and facilitating novel insights into structure-property relationships. Our work presents a significant step toward developing high-performing, chemically interpretable DL models for molecular discovery.}

\keywords{Molecular Property Prediction, Functional Groups, Deep Learning, Interpretability}

\maketitle

\section{Introduction}\label{introduction}
Determining molecule properties is essential in drug, material, and chemical discovery. Typically,  a set of wet laboratory experiments is performed to determine the properties of molecules. This task of molecular property determination is time-consuming and resource-consuming in the discovery process, as several wet laboratory experiments must be carried out. For example, on average, one drug is approved by the US FDA for five compounds entering clinical trials that, in turn, are the result of thorough preclinical testing of 250 compounds themselves selected by screening 5000–10000 compounds~\cite{shen_molecular_2019}. Hence,  computational molecular modelling approaches such as Quantitative Structure-Activity Relationship (QSAR) have been developed to link molecules' physical, chemical, and biological properties with their structure~\cite{walters_applications_2021}. These QSAR strategies allowed chemists to narrow the vast chemical space to a smaller subset of molecules to be synthesised, cutting operational costs and time. However, these approaches relied on limited labelled datasets and hand-crafted features (or molecular representation). In recent years, deep learning-based approaches have been explored to understand complex relations between property and chemical structure based on learned representations instead of relying on expert-curated molecular features~\cite{gilmer_neural_2017, yang_analyzing_2019,walters_applications_2021,fang_geometry-enhanced_2022}. The learned representations can be tailored to specific tasks, leading to a significant increase in prediction performance compared to conventional hand-crafted molecular descriptors and fingerprint features. This instantaneous molecular property prediction using deep learning algorithms can help in different drug and material discovery stages. 

Recently, advances in deep learning approaches, graph, and language-based approaches have resulted in diverse methodologies developed for predicting properties of small molecules ~\cite{yang_analyzing_2019,fang_geometry-enhanced_2022}. The current representation methods for predicting molecular properties can broadly be categorised into four types: (i) Domain knowledge-based representations (fingerprints), (ii) Sequence-based representations, (iii) Graph-based representations, and (iv) Knowledge graph-based representations. Topological fingerprints such as Extended Connectivity Fingerprints (ECFP)~\cite{rogers_extended-connectivity_2010} and Molecular ACCess System (MACCS)~\cite{durant_reoptimization_2002}  based on substructure and molecule similarity search represent molecules as a sequence of bits in an identifier list with each bit indicating the presence or absence of a particular substructure. Kekulescope~\cite{cortes-ciriano_kekulescope_2019} and MolMapNet~\cite{shen_out---box_2021} used deep convolutional neural networks on 2D feature maps of fingerprint features which outperform established models on pharmaceutically relevant benchmarks. This fixed-length binary representation (such as 1024, 2048) typically results in the loss of a certain amount of information, thereby diminishing the quality and interpretability of this representation. Hence,  these fingerprints can hinder the ability to draw meaningful conclusions about structure-activity relationships and make informed molecular design decisions. String representation of molecules, Simplified Molecular-Input Line-Entry System (SMILES)~\cite{weininger_smiles_1988} and Self-Referencing Embedded Strings (SELFIES)~\cite{krenn_self-referencing_2020} was used as input to sequence-based models such as Recurrent Neural Networks and Transformers to learn features automatically for diverse molecular property prediction tasks ~\cite{irwin_chemformer_2022,yuksel_selformer_2023, goh_smiles2vec_2018, ross_large-scale_2022}. Although sequence-based approaches do not capture the inherent molecular structure in the notation, these models can offer interpretable explanations by pinpointing specific chemical components following established knowledge in first-principle chemistry.  

Molecules can be depicted as hydrogen-depleted topological graphs with the atoms as nodes and the bonds between them as edges. Graph Neural Networks (GNN)~\cite{kipf_semi-supervised_2017,xu_how_2018} have been used to learn molecular representations but fail to distinguish between simple structures and are not robust to noise. Message Passing Neural Networks (MPNN) and its variants learn graph-based representations of molecules by conducting sequential message passing to transmit information throughout the molecule using atoms, directed edges~\cite{gilmer_neural_2017,yang_analyzing_2019,song_communicative_2020}. The knowledge of molecular structures can be learned using unsupervised or self-supervised learning strategies from extensive unlabeled molecule data~\cite{liu_n-gram_2019,zhang_motif-based_2024,rong_self-supervised_2020,wang_molecular_2022}. Geometry-Enhanced Molecular (GEM) Representation~\cite{fang_geometry-enhanced_2022}, a spatial learning-based paradigm, accounts for geometries and topology by using an atom-bond graph and a bond-angle graph for learning the representation. Despite the suitability of graph-based representations for molecules and the specific design of GNNs to handle graph-structured data to capture intricate relationships without human intuition, GNNs face certain technical limitations. These include a lack of expressivity~\cite{xu_how_2018} and a limited local receptive field that prevents gathering information from distant atoms. Recently, knowledge graph-enhanced molecular contrastive learning with functional prompt (KANO) has been proposed to bridge the gap between pre-trained and fine-tuned representations by providing a chemical prompt during fine-tuning~\cite{fang_knowledge_2023}. The authors constructed a chemical element-oriented knowledge graph (ElementKG) based on the periodic table and employed an element-guided graph augmentation in contrastive pre-training to understand chemical semantics. The downstream task-related knowledge is retrieved based on prompts generated using the knowledge graph. However, the element-based knowledge graph cannot capture molecular system complexity, and the functional prompts might not capture long-range dependencies between substructures.

Although Graph Neural Networks (GNNs) and self-supervised learning models (language models) have shown promise in property prediction tasks~\cite{kipf_semi-supervised_2017,xu_how_2018,goh_smiles2vec_2018,irwin_chemformer_2022,wang_molecular_2022,yuksel_selformer_2023}, interpreting the relationship between properties and molecule structures remains challenging. This difficulty stems from the complex molecular representations these methods generate, obtained by pre-training on massive datasets. Chemists need help deciphering these intricate representations, hindering their ability to gain chemical intuition from the models. For novel molecule discovery and drug repurposing applications, chemically interpretable molecular representation is essential for testing the generated molecules via wet lab experiments by chemists. Introducing interpretability to features and models results in more effective training, improved generalisation, and reduced occurrence of adversarial examples~\cite{rudin_stop_2019}. Ensuring the interpretability of features guarantees that our model utilises relevant information for our target, thereby lowering the risk of the model capturing spurious correlations. On the other hand, molecular fingerprints provide a straightforward and interpretable representation of molecular structures and encode molecular features into binary vectors, making them easy to understand and allowing the direct examination of the presence or absence of specific molecular features associated with predicted properties. Hence, a chemistry-inspired representation of molecules can be vital to achieving interpretability and enhanced prediction performance for these models. 

In this work, we propose a molecular representation learning framework that uses the concept of functional groups in chemistry. The functional groups are substructures in a molecule attributed to its chemical properties and reactivity. This work proposes a functional group representation (FGR) framework that allows embedding molecules based on their substructures.
To the best of our knowledge, this work is the first attempt to incorporate the concept of functional groups in chemistry using interpretable structural keys relevant to molecular property prediction tasks. Notably, our model boasts superior efficiency in terms of parameter count and architecture simplicity compared to existing methodologies. This streamlined design enhances computational efficiency and facilitates model interpretability, allowing for more precise insights into the underlying chemical representations learned by the model.

The paper is organised as follows: Firstly, we introduce two approaches for the generation of the functional group vocabulary, namely, functional groups (FG) curated from established chemistry publications and mined functional groups (MFG) from the PubChem database~\cite{kim_pubchem_2016}. Including functional groups as input features adheres to properties intrinsically linked to interpretability elements, including readability, understandability, and relevancy~\cite{zytek_need_2022}. This alignment facilitates a heightened level of trust among chemists, thereby increasing the likelihood of their utilisation in practical scenarios. We perform experiments on several benchmark datasets in the available literature and compare the results of the proposed FGR framework in this work with other state-of-the-art methods. We demonstrate that the FGR framework outperforms several property prediction tasks and provides comparable results on several other tasks compared to the state-of-the-art methods while providing chemical interpretability to chemists and practitioners. We verify the interpretability of the models using literature-reported functional groups for different datasets.   

\begin{figure}[!htbp]
    \centering
    \includegraphics[width=0.7\textwidth]{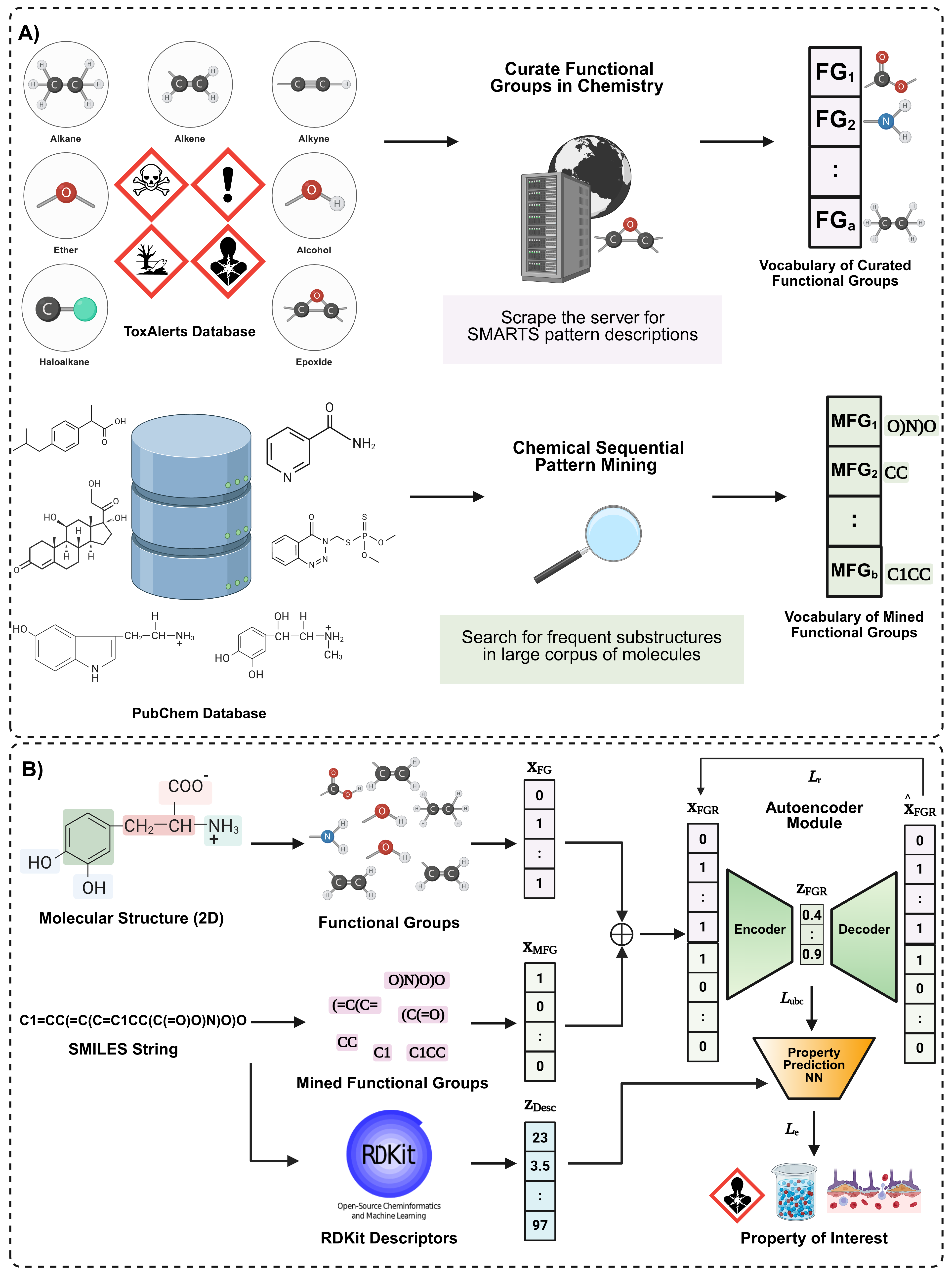}
    \caption{\textbf{A)} Generation of functional group vocabulary for curated Functional Groups (FG) and Mined Functional Groups (MFG) \textbf{B)} Latent Feature Embedding for FG representation, MFG representation along with the combined representation (FGR)}
    \label{fig:introduction_methodology}
\end{figure}

\section{Results}\label{results}

\subsection{Overview of Functional Group Representation Framework}

The proposed chemistry-inspired framework for learning molecular representation using functional groups (or constituent substructures) is termed Functional Group Representation (FGR). The problem setting is described in~\cref{problem_setting}. The FGR framework consists of two steps: 
\begin{enumerate}
    \item Generation of functional group vocabulary for multi-one hot encoding as shown in~\cref{fig:introduction_methodology} (A). In this step, PubChem and ToxAlerts Databases are used to generate functional group vocabulary. A sequential pattern mining algorithm generates a vocabulary of the mined functional groups using SMILES in PubChem Database. The vocabulary of functional groups is generated by scraping the curated functional groups from the ToxAlerts database. Details on the curation of functional groups and the pattern mining algorithm are provided in~\cref{fg_vocab}.
      
    \item In the second step, latent feature embedding of molecules using functional groups vocabulary (FG and MFG) generated in the previous step using autoencoders as shown in~\cref{fig:introduction_methodology} (B). The latent feature embedding of molecules with the molecular descriptors is then used for different downstream property prediction tasks. More details on the latent feature embedding task are given in~\cref{latent_feature}. 
\end{enumerate}
The model is trained end-to-end, combining latent feature embedding and property prediction using a feedforward neural network. More details on the outputs and loss functions are provided in~\cref{prediction_task}.

\subsection{Functional Groups-Inspired Representation (FGR) achieves State-of-the-Art (SOTA) Performance in Molecular Property Prediction}
To assess the performance of our framework, we rigorously evaluated its performance on a comprehensive range of datasets in seven categories: physiology, biophysics, physical chemistry, quantum mechanics, bioactivity, pharmacokinetics, and cleavage of proteins using peptides. The molecular properties of the datasets are varied and encompass a broad range of characteristics. These characteristics include but are not limited to the ability to penetrate the blood-brain barrier, electronic properties, inhibition of $\beta$-Secretase 1 enzyme, inhibition of cancer cell line growth, liver microsomal clearance, and cleavage of SARS-CoV-2 main protease. For more information on the datasets, refer to Supplementary Information S1. All the results presented in this work are based on a scaffold split, which ensures that molecules in the test set have distinct core structures from those in the training set. This approach provides a more rigorous evaluation of model generalization to structurally novel compounds. The tables mention the number of molecules and the number of binary prediction tasks (multi-task), along with SOTA results highlighted in bold and underlined, indicating the second-best performing model. For detailed information on the dataset split, baselines used to compare our framework, training and performance evaluation refer to~\cref{experimental_setup}.

\subsubsection{MoleculeNet Datasets}
\cref{tab:classification_scaffold,tab:regression_scaffold} summarize results from the latest SOTA methods, including the proposed FGR using MoleculeNet datasets.~\cref{tab:classification_scaffold} presents the mean and standard deviation of test ROC-AUC (\%) on three independent runs of physiology and biophysics datasets. The proposed FGR approach outperforms the current SOTA method KANO~\cite{fang_knowledge_2023} on five tasks out of eight, showing a 1.47\% improvement over KANO. Additionally, FGR achieves the second-highest score in one task (BACE). The combined representation achieves top scores among self-supervised, graph-based, and other supervised learning methods, indicating that the FGR representation can capture varying levels of molecular complexity. The representation that combines well-curated structural keys from the ToxAlerts database~\cite{sushko_toxalerts_2012} with evident mechanisms of action performs well on toxicity-related datasets.

\cref{tab:regression_scaffold} presents the mean and standard deviation of test Root Mean Squared Error (RMSE) (for ESOL, FreeSolv, and Lipophilicity) or mean absolute error (qm7, qm8, and qm9) on three independent runs. Our model achieves SOTA performance in two of three physical chemistry tasks and comparable performance in quantum mechanics tasks. The average improvement over the physical chemistry tasks is observed to be 8.66\%. The framework performs well in datasets with fewer labelled molecules, even without pre-training. The representation performs well in the physical chemistry datasets, suggesting that incorporating functional group patterns, such as hydroxyl and amino groups for hydrophilic properties and alkyl and phenyl groups for hydrophobic properties, is beneficial.

The framework shows limitations in datasets like HIV and MUV, where the challenges of imbalanced data and the exclusion of 3D geometries are prominent. The absence of 3D molecular information may affect performance for quantum mechanics tasks, as these properties are closely tied to molecular geometry and element-level composition. More details on label distribution can be found in Supplementary Information S1.

\begin{table}[!htbp]
\centering
\resizebox{\textwidth}{!}{%
\begin{tabular}{|c|c|c|c|c|c|c|c|c|}
\hline
\multicolumn{1}{|c|}{\textbf{Category}} & \multicolumn{5}{|c|}{\textbf{Physiology}} & \multicolumn{3}{|c|}{\textbf{Biophysics}}\\
\hline
\multicolumn{1}{|c|}{\textbf{Dataset}} & \textbf{BBBP $\uparrow$} & \textbf{Tox21 $\uparrow$} & \textbf{ToxCast $\uparrow$} & \textbf{SIDER $\uparrow$} & \textbf{ClinTox $\uparrow$} & \textbf{BACE $\uparrow$} & \textbf{MUV $\uparrow$} & \textbf{HIV $\uparrow$}\\
\hline
\multicolumn{1}{|c|}{\textbf{Molecules}} & \textbf{2,039} & \textbf{7,831} & \textbf{8,575} & \textbf{1,427} & \textbf{1,478} & \textbf{1,513} & \textbf{93,807} & \textbf{41,127}\\
\hline
\multicolumn{1}{|c|}{\textbf{Tasks}} & \textbf{1} & \textbf{12} & \textbf{617} & \textbf{27} & \textbf{2} & \textbf{1} & \textbf{17} & \textbf{1}\\
\hline

GCN~\cite{kipf_semi-supervised_2017} & $71.8 \pm 0.9$ & $70.9 \pm 0.3$ & $65.0 \pm 6.1$ & $53.6 \pm 0.3$ & $62.5 \pm 2.8$ & $71.6 \pm 2.0$ & $71.6 \pm 4.0$ & $74.0 \pm 3.0$ \\\hline
MPNN~\cite{gilmer_neural_2017} & $91.3 \pm 4.1$ & $80.8 \pm 2.4$ & $69.1 \pm 3.0$ & $59.5 \pm 3.0$ & $87.9 \pm 5.4$ & $81.5 \pm 1.0$ & $75.7 \pm 1.3$ & $77.0 \pm 1.4$ \\\hline
GIN~\cite{xu_how_2018} & $65.8 \pm 4.5$ & $74.0 \pm 0.8$ & $66.7 \pm 1.5$ & $57.3 \pm 1.6$ & $58.0 \pm 4.4$ & $70.1 \pm 5.4$ & $71.8 \pm 2.5$ & $75.3 \pm 1.9$ \\\hline
N-GRAM~\cite{liu_n-gram_2019} & $91.2 \pm 0.3$ & $76.9 \pm 2.7$ & - & $63.2 \pm 0.5$ & $87.5 \pm 2.7$ & $79.1 \pm 1.3$ & $76.9 \pm 0.7$ & $78.7 \pm 0.4$ \\\hline
DMPNN~\cite{yang_analyzing_2019} & $91.9 \pm 3.0$ & $75.9 \pm 0.7$ & $63.7 \pm 0.2$ & $57.0 \pm 0.7$ & $90.6 \pm 0.6$ & $85.2 \pm 0.6$ & $78.6 \pm 1.4$ & $77.1 \pm 0.5$ \\\hline
CMPNN~\cite{song_communicative_2020} & $92.7 \pm 1.7$ & $80.1 \pm 1.6$ & $70.8 \pm 1.3$ & $61.6 \pm 0.3$ & $89.8 \pm 0.8$ & $86.7 \pm 0.2$ & $\CP \underline{79.0 \pm 2.0}$ & $78.2 \pm 2.2$ \\\hline
GROVER~\cite{rong_self-supervised_2020} & $86.8 \pm 2.2$ & $80.3 \pm 2.0$ & $56.8 \pm 3.4$ & $61.2 \pm 2.5$ & $70.3 \pm 13.7$ & $82.4 \pm 3.6$ & $67.3 \pm 1.8$ & $68.2 \pm 1.1$ \\\hline
MGSSL~\cite{zhang_motif-based_2024} & $70.5 \pm 1.1$ & $76.4 \pm 0.4$ & $64.1 \pm 0.7$ & $61.8 \pm 0.8$ & $80.7 \pm 2.1$ & $79.7 \pm 0.8$ & $78.7 \pm 1.5$ & $79.5 \pm 1.1$ \\\hline
GEM~\cite{fang_geometry-enhanced_2022} & $88.8 \pm 0.4$ & $78.1 \pm 0.4$ & $68.6 \pm 0.2$ & $63.2 \pm 1.5$ & $90.3 \pm 0.7$ & $87.9 \pm 1.1$ & $75.3 \pm 1.5$ & $\CP \underline{81.3 \pm 0.3}$ \\\hline
GraphMVP~\cite{liu_pre-training_2021} & $72.4 \pm 1.6$ & $75.9 \pm 0.5$ & $63.1 \pm 0.4$ & $63.9 \pm 1.2$ & $79.1 \pm 2.8$ & $81.2 \pm 0.9$ & $77.7 \pm 0.6$ & $77.0 \pm 1.2$ \\\hline
MolCLR~\cite{wang_molecular_2022} & $73.3 \pm 1.0$ & $74.1 \pm 5.3$ & $65.9 \pm 2.1$ & $61.2 \pm 3.6$ & $89.8 \pm 2.7$ & $82.8 \pm 0.7$ & $78.9 \pm 2.3$ & $77.4 \pm 0.6$ \\\hline
MolCLR$_{\mathrm{CMPNN}}$ & $72.4 \pm 0.7$ & $78.4 \pm 2.6$ & $69.1 \pm 1.2$ & $59.7 \pm 3.4$ & $88.0 \pm 4.0$ & $85.0 \pm 2.4$ & $74.5 \pm 2.1$ & $77.8 \pm 5.5$ \\\hline
KANO~\cite{fang_knowledge_2023} & $\CC \mathbf{96.0 \pm 1.6}$ & $\CP \underline{83.7 \pm 1.3}$ & $\CP \underline{73.2 \pm 1.6}$ & $\CP \underline{65.2 \pm 0.8}$ & $\CP \underline{94.4 \pm 0.3}$ & $\CC \mathbf{93.1 \pm 2.1}$ & $\CC \mathbf{83.7 \pm 2.3}$ & $\CC \mathbf{85.1 \pm 2.2}$ \\ \hline\hline
FG$^\#$ & $\CP \underline{93.9 \pm 2.8}$ & $78.3 \pm 1.2$ & $72.7 \pm 0.6$ & $60.5 \pm 1.7$ & $88.8 \pm 7.6$ & $87.2 \pm 2.4$ & $72.3 \pm 1.3$ & $77.5 \pm 2.7$ \\ \hline
MFG$^\#$ & $88.4 \pm 1.5$ & $67.5 \pm 0.7$ & $63.0 \pm 1.2$ & $56.0 \pm 0.7$ & $69.7 \pm 5.4$ & $84.9 \pm 4.2$ & $68.6 \pm 2.5$ & $76.7 \pm 4.8$ \\ \hline
FGR$^\#$ & $\CC \mathbf{96.0 \pm 1.8}$ & $\CC \mathbf{84.1 \pm 1.0}$ & $\CC \mathbf{74.0 \pm 2.1}$ & $\CC \mathbf{67.8 \pm 2.8}$ & $\CC \mathbf{96.1 \pm 0.5}$ & $\CP \underline{89.3 \pm 3.1}$ & $74.2 \pm 4.1$ & $78.3 \pm 1.1$ \\ \hline
\end{tabular}%
}
\caption{The mean and standard deviation of test ROC-AUC (\%) on three independent runs are reported. The dataset split is based on molecular scaffolds. \colorbox{aliceblue}{\textbf{Bold}} indicates the best performing model and \colorbox{babypink}{\underline{underline}} indicates the second best performing model. $\#$ indicates the concatenation of descriptors.}
\label{tab:classification_scaffold}
\end{table}

\begin{table}[!htbp]
\centering
\resizebox{\textwidth}{!}{%
\begin{tabular}{|c|c|c|c|c|c|c|}
\hline
\multicolumn{1}{|c|}{\textbf{Category}} & \multicolumn{3}{|c|}{\textbf{Physical Chemistry}} & \multicolumn{3}{|c|}{\textbf{Quantum Mechanics}}\\
\hline
\multicolumn{1}{|c|}{\textbf{Dataset}} & \textbf{ESOL $\downarrow$} & \textbf{FreeSolv $\downarrow$} & \textbf{Lipophilicity $\downarrow$} & \textbf{qm7 $\downarrow$} & \textbf{qm8 $\downarrow$} & \textbf{qm9\tablefootnote{We consider only the “homo,” “lumo,” and “gap” targets from the QM9 dataset, as the remaining targets exhibit significantly different value ranges. The average mean absolute error (MAE) is then computed over these three selected properties.} $\downarrow$} \\
\hline
\multicolumn{1}{|c|}{\textbf{Molecules}} & \textbf{1,128} & \textbf{642} & \textbf{4,200} & \textbf{7,160} & \textbf{21,786} & \textbf{133,885}\\
\hline
\multicolumn{1}{|c|}{\textbf{Tasks}} & \textbf{1} & \textbf{1} & \textbf{1} & \textbf{12} & \textbf{3}& \textbf{3}\\
\hline

GCN~\cite{kipf_semi-supervised_2017} & $1.431 \pm 0.050$ & $2.870 \pm 0.135$ & $0.712 \pm 0.049$ & $122.9 \pm 2.2$ & $0.0366 \pm 0.000$ & $0.00835 \pm 0.00001$ \\\hline
MPNN~\cite{gilmer_neural_2017} & $1.167 \pm 0.430$ & $1.621 \pm 0.952$ & $0.672 \pm 0.051$ & $111.4 \pm 0.9$ & $\CP \underline{0.0148 \pm 0.001}$ & $0.00522 \pm 0.00003$ \\\hline
GIN~\cite{xu_how_2018} & $1.452 \pm 0.020$ & $2.765 \pm 0.180$ & $0.850 \pm 0.071$ & $124.8 \pm 0.7$ & $0.0371 \pm 0.001$ & $0.00824 \pm 0.00004$ \\\hline
N-GRAM~\cite{liu_n-gram_2019} & $1.100 \pm 0.030$ & $2.510 \pm 0.191$ & $0.880 \pm 0.121$ & $125.6 \pm 1.5$ & $0.0320 \pm 0.003$ & $0.00964 \pm 0.00031$ \\\hline
DMPNN~\cite{yang_analyzing_2019} & $1.050 \pm 0.008$ & $1.673 \pm 0.082$ & $0.683 \pm 0.016$ & $103.5 \pm 8.6$ & $0.0156 \pm 0.001$ & $0.00514 \pm 0.00001$ \\\hline
CMPNN~\cite{song_communicative_2020} & $0.798 \pm 0.112$ & $1.570 \pm 0.442$ & $\CP \underline{0.614 \pm 0.029}$ & $75.1 \pm 3.1$ & $0.0153 \pm 0.002$ & $\CP \underline{0.00405 \pm 0.00002}$ \\\hline
GROVER~\cite{rong_self-supervised_2020} & $1.423 \pm 0.288$ & $2.947 \pm 0.615$ & $0.823 \pm 0.010$ & $91.3 \pm 1.9$ & $0.0182 \pm 0.001$ & $0.00719 \pm 0.00208$ \\\hline
GEM~\cite{fang_geometry-enhanced_2022} & $0.813 \pm 0.028$ & $1.748 \pm 0.114$ & $0.674 \pm 0.022$ & $60.0 \pm 2.7$ & $0.0163 \pm 0.001$ & $0.00562 \pm 0.00007$ \\\hline
MolCLR~\cite{wang_molecular_2022} & $1.113 \pm 0.023$ & $2.301 \pm 0.247$ & $0.789 \pm 0.009$ & $90.9 \pm 1.7$ & $0.0185 \pm 0.013$ & $0.00480 \pm 0.00003$ \\\hline
MolCLR$_{\mathrm{CMPNN}}$ & $0.911 \pm 0.082$ & $2.021 \pm 0.133$ & $0.875 \pm 0.003$ & $89.8 \pm 6.3$ & $0.0179 \pm 0.001$ & $0.00475 \pm 0.00001$ \\\hline
KANO~\cite{fang_knowledge_2023} & $\CP \underline{0.670 \pm 0.019}$ & $1.142 \pm 0.258$ & $\CC \mathbf{0.566 \pm 0.007}$ & $\CP \underline{56.4 \pm 2.8}$ & $\CC \mathbf{0.0123 \pm 0.000}$ & $\CC \mathbf{0.00320 \pm 0.00001}$ \\ \hline\hline
FG$^\#$ & $0.763 \pm 0.071$ & $\CP \underline{0.825 \pm 0.221}$ & $0.742 \pm 0.050$ & $59.2 \pm 1.7$ & $0.0335 \pm 0.003$ & $0.00690 \pm 0.00005$ \\ \hline
MFG$^\#$ & $0.812 \pm 0.083$ & $1.034 \pm 0.100$ & $0.757 \pm 0.025$ & $61.6 \pm 1.9$ & $0.0351 \pm 0.003$ & $0.00730 \pm 0.00010$ \\ \hline
FGR$^\#$ & $\CC \mathbf{0.620 \pm 0.067}$ & $\CC \mathbf{0.789 \pm 0.192}$ & $0.636 \pm 0.027$ &$\CC \mathbf{55.3 \pm 1.6}$ & $0.0297 \pm 0.003$ & $0.00547 \pm 0.00008$ \\ \hline
\end{tabular}%
}
\caption{The mean and standard deviation of test root mean square error (for ESOL, FreeSolv and Lipophilicity) or mean absolute error (for qm7, qm8 and qm9) on three independent runs are reported. The dataset split is based on molecular scaffolds. \colorbox{aliceblue}{\textbf{Bold}} indicates the best performing model and \colorbox{babypink}{\underline{underline}} indicates the second best performing model. $\#$ indicates the concatenation of descriptors.}
\label{tab:regression_scaffold}
\end{table}

\subsubsection{MolMapNet Datasets}

\cref{tab:bioactivity_scaffold,tab:pharmaco_scaffold} summarize results from the latest SOTA methods using MolMapNet datasets. \cref{tab:bioactivity_scaffold} presents the mean of test $\mathrm{R}^2$ (for cancer cell lines), or RMSE (for Malaria) on three independent runs. The cancer cell lines dataset investigates the effect of chemicals on different biological targets quantified using pIC$_{50}$. The combined representation achieves the highest scores among CCRF-CEM, KB, LoVO, PC-3, SK-OV-3, and Malaria datasets, improving over previous SOTA methods Kekulescope and MolMapNet. \cref{tab:pharmaco_scaffold} presents the mean of test $\mathrm{R}^2$ (for LMC) or ROC-AUC (for CYP) on three independent runs. The combined representation beats SOTA methods on nine out of fourteen MolMapNet datasets, yielding an overall improvement of 2.3\%.

\begin{table}[ht]
\centering
\resizebox{\textwidth}{!}{%
\begin{tabular}{|c|c|c|c|c|c|c|c|c|c|}
\hline
\multicolumn{1}{|c|}{\textbf{Category}} & \multicolumn{9}{|c|}{\textbf{Bioactivity}}\\
\hline
\multicolumn{1}{|c|}{\textbf{Dataset}} & \textbf{A2780 $\uparrow$} & \textbf{CCRF-CEM $\uparrow$} & \textbf{DU-145 $\uparrow$} & \textbf{HCT-15 $\uparrow$} & \textbf{KB $\uparrow$} & \textbf{LoVo $\uparrow$} & \textbf{PC-3 $\uparrow$} & \textbf{SK-OV-3 $\uparrow$} & \textbf{Malaria $\downarrow$}\\
\hline
\multicolumn{1}{|c|}{\textbf{Molecules}} & \textbf{2,255} & \textbf{3,047} & \textbf{2,512} & \textbf{994} & \textbf{2,731} & \textbf{1,120} & \textbf{4,294} & \textbf{1,589} & \textbf{9,998}\\ 
\hline
\multicolumn{1}{|c|}{\textbf{Tasks}} & \textbf{1} & \textbf{1} & \textbf{1} & \textbf{1} & \textbf{1}& \textbf{1} & \textbf{1} & \textbf{1} & \textbf{1}\\
\hline

Kekulescope~\cite{cortes-ciriano_kekulescope_2019} & $0.622$ & $0.528$ & $0.427$ & $\CP \underline{0.617}$ & $0.533$ & $0.530$ & $0.496$ & $0.461$ & -\\\hline
MolMapNet~\cite{shen_out---box_2021} & $\CC \mathbf{0.663}$ & $0.627$ & $\CC \mathbf{0.594}$ & $\CC \mathbf{0.734}$ & $\CC \mathbf{0.713}$ & $\CP \underline{0.583}$ & $\CP \underline{0.615}$ & $\CP \underline{0.597}$ & $1.011$ \\\hline\hline
FG$^\#$ & $0.624$ & $\CP \underline{0.642}$ & $0.540$ & $0.529$ & $0.618$ & $0.577$ & $0.496$ & $0.561$ & $\CP \underline{0.981}$\\\hline
MFG$^\#$ & $0.597$ & $0.611$ & $0.357$ & $0.593$ & $0.516$ & $0.523$ & $0.472$ & $0.385$ & $1.156$\\\hline
FGR$^\#$ & $\CP \underline{0.632}$ & $\CC \mathbf{0.662}$ & $\CP \underline{0.563}$ & $0.607$ & $\CP \underline{0.627}$ & $\CC \mathbf{0.619}$ & $\CC \mathbf{0.639}$ & $\CC \mathbf{0.627}$ & $\CC \mathbf{0.938}$\\\hline
\end{tabular}%
}
\caption{The mean of test $\mathrm{R}^2$ (for cancer cell lines) or RMSE (for Malaria) on three independent runs is reported. The dataset split is based on molecular scaffolds. \colorbox{aliceblue}{\textbf{Bold}} indicates the best performing model and \colorbox{babypink}{\underline{underline}} indicates the second best performing model. $\#$ indicates the concatenation of descriptors.}
\label{tab:bioactivity_scaffold}
\end{table}

\begin{table}[ht]
\centering
\resizebox{0.7\textwidth}{!}{%
\begin{tabular}{|c|c|c|c|c|}
\hline
\multicolumn{1}{|c|}{\textbf{Category}} & \multicolumn{4}{|c|}{\textbf{Pharmacokinetic}}\\
\hline
\multicolumn{1}{|c|}{\textbf{Dataset}} & \textbf{CYP $\uparrow$} & \textbf{LMC-H $\uparrow$} & \textbf{LMC-R $\uparrow$} & \textbf{LMC-M $\uparrow$}\\
\hline
\multicolumn{1}{|c|}{\textbf{Molecules}} & \textbf{16,896} & \textbf{8,755} & \textbf{8,755} & \textbf{8,755}\\ 
\hline
\multicolumn{1}{|c|}{\textbf{Tasks}} & \textbf{5} & \textbf{1} & \textbf{1} & \textbf{1}\\
\hline

Kekulescope~\cite{cortes-ciriano_kekulescope_2019} & $88.4$ & $0.566$ & $0.771$ & $0.475$ \\\hline
MolMapNet~\cite{shen_out---box_2021} & $\CP \underline{88.6}$ & $\CP \underline{0.580}$ & $\CP \underline{0.790}$ & $0.526$ \\\hline\hline
FG$^\#$ & $87.9$ & $0.551$ & $0.783$ & $0.548$\\\hline
MFG$^\#$ & $79.8$ & $0.539$ & $0.736$ & $\CP \underline{0.553}$\\\hline
FGR$^\#$ & $\CC \mathbf{92.3}$ & $\CC \mathbf{0.623}$ & $\CC \mathbf{0.814}$ & $\CC \mathbf{0.578}$ \\\hline
\end{tabular}%
}
\caption{The mean of test $\mathrm{R}^2$ (for LMC) or ROC-AUC (for CYP) on three independent runs are reported. The dataset split is based on molecular scaffolds. \colorbox{aliceblue}{\textbf{Bold}} indicates the best performing model and \colorbox{babypink}{\underline{underline}} indicates the second best performing model. $\#$ indicates the concatenation of descriptors.}
\label{tab:pharmaco_scaffold}
\end{table}

\subsection{Peptide Cleavage and Bacterial Datasets}
\cref{tab:peptide_scaffold} presents the mean and standard deviation of test ROC-AUC (\%) on three independent runs for peptide cleavage and antibiotic activity datasets. We compare the FGR framework with DMPNN~\cite{yang_analyzing_2019}, a graph-based SOTA method for molecular property prediction in bacterial and viral benchmark datasets. Our framework beat the SOTA method with a 1.94\% average margin on all the datasets. The graph method is limited to capturing local dependencies. Hence, the DMPNN may not be scalable for datasets containing large molecules, as in the case of peptides. In contrast, our framework, containing a fixed input size, can scale to any arbitrary molecule size. The length distribution of the SMILES strings is available in Supplementary Information S1. 

\begin{table}[ht]
\centering
\resizebox{\textwidth}{!}{%
\begin{tabular}{|c|c|c|c|c|c|c|}
\hline
\multicolumn{1}{|c|}{\textbf{Category}} & \multicolumn{6}{|c|}{\textbf{Peptide Cleavage}}\\
\hline
\multicolumn{1}{|c|}{\textbf{Dataset}} & \textbf{746\_aa $\uparrow$} & \textbf{1625\_aa $\uparrow$} & \textbf{Schilling $\uparrow$} & \textbf{Impens $\uparrow$} & \textbf{Mpro $\uparrow$} & \textbf{\textit{E.~coli} $\uparrow$}\\
\hline
\multicolumn{1}{|c|}{\textbf{Molecules}} & \textbf{746} & \textbf{1,625} & \textbf{3272} & \textbf{947} & \textbf{880} & \textbf{2335}\\ 
\hline
\multicolumn{1}{|c|}{\textbf{Tasks}} & \textbf{1} & \textbf{1} & \textbf{1} & \textbf{1} & \textbf{1} & \textbf{1}\\
\hline

DMPNN~\cite{yang_analyzing_2019} & $94.2\pm3.4$ & $\CP \underline{98.1\pm1.6}$ & $\CP \underline{95.6\pm2.9}$ & $\CP \underline{86.7\pm2.5}$ & $\CP \underline{77.3\pm9.6}$ & $\CP \underline{89.0\pm5.4}$\\\hline\hline
FG$^\#$ & $89.1 \pm 6.4$ & $97.2\pm2.1$ & $92.5\pm3.2$ & $81.7\pm3.4$ & $74.1\pm9.2$ & $85.9\pm5.9$\\\hline
MFG$^\#$ & $\CP \underline{96.5 \pm 1.0}$ & $95.6\pm2.3$ & $91.1\pm3.6$ & $80.0\pm5.7$ & $73.5\pm9.7$ & $85.7\pm6.1$\\\hline
FGR$^\#$ & $\CC \mathbf{97.9 \pm 1.3}$ & $\CC \mathbf{98.9\pm0.7}$ & $\CC \mathbf{96.5\pm2.2}$ & $\CC \mathbf{89.3\pm2.7}$ & $\CC \mathbf{80.9\pm9.3}$ & $\CC \mathbf{93.5\pm5.6}$\\\hline
\end{tabular}%
}
\caption{The mean and standard deviation of test ROC-AUC (\%) on three independent runs are reported. The dataset split is based on molecular scaffolds. \colorbox{aliceblue}{\textbf{Bold}} indicates the best performing model and \colorbox{babypink}{\underline{underline}} indicates the second best performing model. $\#$ indicates the concatenation of descriptors.}
\label{tab:peptide_scaffold}
\end{table}

The combination of FGR encoding consistently outperforms the individual FG and MFG encodings, demonstrating the strength of integrating both approaches. The combined encoding captures a broader range of molecular features by leveraging functional group patterns curated from databases (FG) alongside mined functional groups identified through pattern mining in SMILES strings (MFG). This complementary nature of FG and MFG enables a more comprehensive molecular representation, leading to improved property prediction performance. These findings highlight the importance of utilizing curated and mined structural keys for accurate and robust molecular representation learning. Additional ablation studies on individual representations are presented in Supplementary Information S2, while details regarding the pre-training procedure of the autoencoder are available in Supplementary Information S3. Results pertaining to the cluster-based dataset split are reported in Supplementary Information S4.

\subsection{Quality of Functional Group Feature Space}
Assessing the quality of the feature space in deep learning models is essential for understanding model behaviour and performance. Two key analyses, alignment and uniformity, provide valuable insights into how features are distributed and organized within a dataset. Alignment analysis would reveal how well the model groups molecules with similar chemical functionalities. Molecules containing the same or chemically similar functional groups should exhibit high alignment, indicating that the representation space correctly captures chemical similarity relationships. Uniformity analysis becomes particularly valuable for molecular representations because it addresses a critical challenge in chemical machine learning: ensuring adequate coverage of chemical space. Poor uniformity would indicate that certain regions of chemical space are over-represented while others remain sparsely populated, potentially leading to biased property predictions. By applying alignment and uniformity metrics to molecular representations, one obtains quantitative measures of representation quality that correlate directly with task performance, particularly in scenarios where minor variations in functional groups give rise to substantially different molecular properties.

\begin{figure}[htb]
    \centering
    \includegraphics[width=\textwidth]{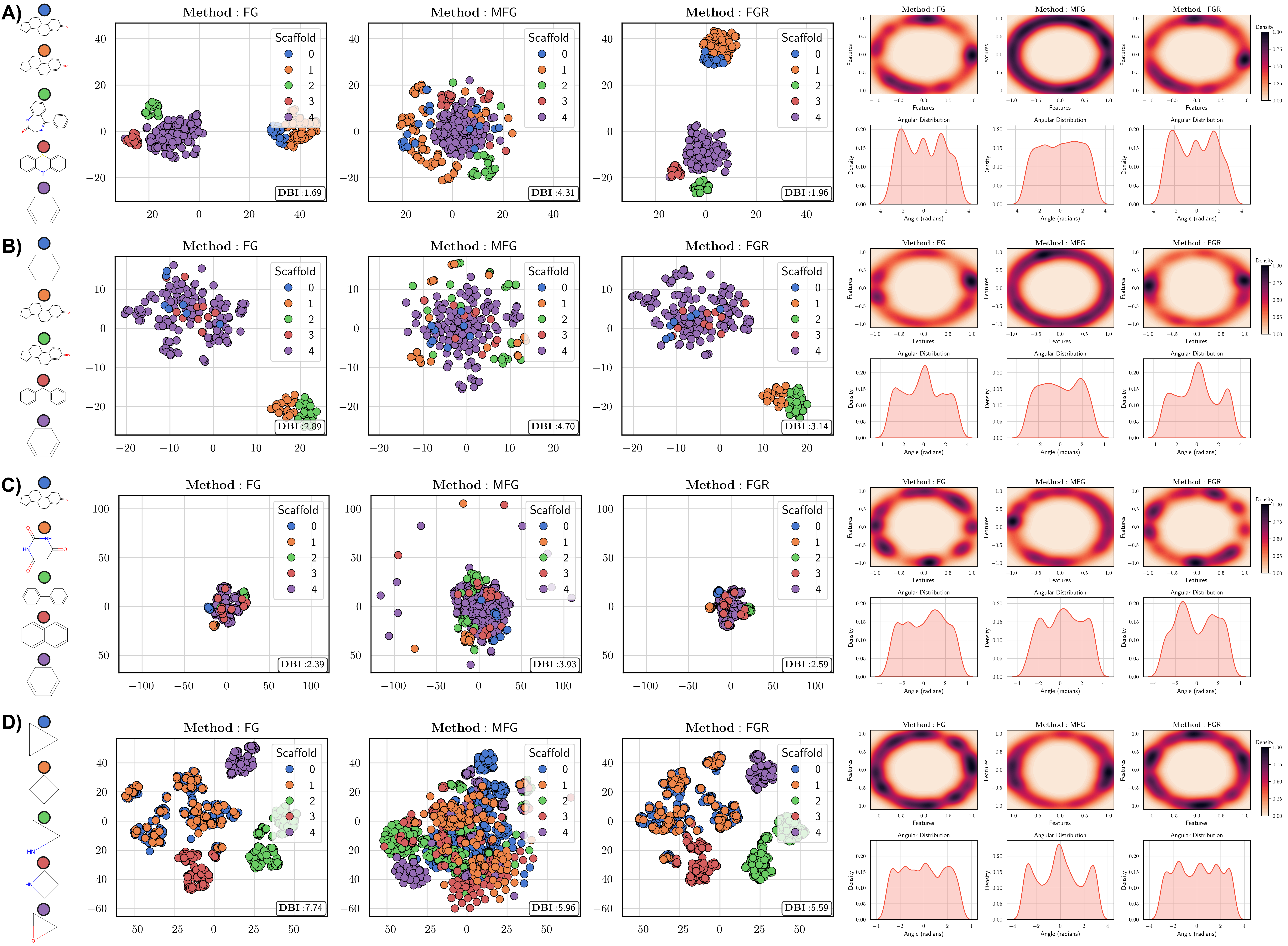}
    \caption{Alignment and Uniformity analysis for \textbf{A)} BBBP \textbf{B)} ClinTox \textbf{C)} ESOL and \textbf{D)} qm7 datasets. \\
    \textbf{Alignment analysis}: The t-SNE visualizations of the input representations indicate the separation of molecules based on dissimilar scaffolds. Different colours indicate distinct scaffolds. Lower DBI indicates better separation of clusters. \\
    \textbf{Uniformity analysis}: Darker regions in the feature density curve indicate more concentration of data points, and flatter curves in the density estimation curve of angles indicate a more uniform distribution.}
    \label{fig:alignment_uniformity}
\end{figure}

\subsubsection{Alignment Analysis}
Alignment analysis of input representations helps to ensure that representations capture relevant information effectively and aid in model performance improvement. It involves assessing the degree of similarity between different input data representations, such as word embeddings, image features, or numerical vectors. We visualize the representations of molecules in $\mathbb{R}^2$ with different scaffolds using t-distributed stochastic neighbour embedding (t-SNE)~\cite{maaten_visualizing_2008}. The ideal representation method should be able to produce distinct clusters with molecules containing the same scaffold to be grouped. We use the Davies Bouldin Index (DBI)~\cite{davies_cluster_1979} to evaluate the clustering quality with a lower DBI indicating better separation of clusters. We chose the top five scaffolds from each dataset, where different colours indicate distinct scaffolds. We also perform the alignment analysis to evaluate the degree of separation between the labels of datasets across different methods (FG, MFG, FGR).

As indicated in~\cref{fig:alignment_uniformity} across the BBBP, Clintox, and ESOL datasets, the FG representation achieved the consistently lowest DBI, indicating the most well-separated clusters in the aligned representations. The FGR representation closely followed FG in the four datasets, demonstrating strong cluster separation. However, the MFG representation method yielded the highest DBI in all datasets. The high DBI scores suggest that MFG might generate less well-defined clusters in the aligned representations. Interestingly, in a single exception on the qm7 dataset, FGR achieved the lowest DBI, highlighting a potential dataset-specific advantage for this combined approach in representation.

This trend holds across the remaining datasets as well, and in each case, FG and FGR outperformed MFG in terms of producing compact and distinct clusters. Further alignment analyses are presented in Supplementary Information S5.

\subsubsection{Uniformity Analysis}
To perform uniformity analysis, we map the input representations onto a unit hypersphere $\mathcal{S}^1$ using t-SNE and visualize in $\mathbb{R}^2$ using a Gaussian kernel density estimator to estimate the density distribution of the projected features on the hypersphere. We divide each feature vector by its Euclidean norm to ensure it lies on the unit hypersphere. The normalization projects the data onto a surface where all points are equidistant from the origin, allowing for equal representation of features. After selecting an appropriate bandwidth parameter (bw=0.2), a smooth representation of the feature density is created, highlighting regions of high and low concentration of data points. The density estimations of angles for each point ($\mathrm{arctan2}(y,x) \forall (x,y) \in \mathcal{S}^1$) are also shown for clarity.

Based on observations from~\cref{fig:alignment_uniformity}, MFG has the most evenly distributed features, whereas FG exhibited sharper peaks, indicating a higher concentration of data points in specific value ranges. Combining FG and MFG in FGR resulted in a distribution that balanced these extremes, reducing the sharpness observed in FG. Consistent trends are also observed across the remaining datasets, as shown in Supplementary Information S6, reinforcing the generalizability of these distributional characteristics.

Our analysis revealed an interesting trade-off between alignment and uniformity. While MFG achieved the most uniform feature distribution, it resulted in the poorest alignment of representations. Conversely, FG excelled in alignment but exhibited the least uniform feature distribution. The combined representation (FGR) strikes a balance between these two aspects. By incorporating elements of both methods, FGR achieves a mid-range level of uniformity while maintaining strong alignment, suggesting it may be the optimal choice for our task of property prediction.

\section{Interpretability Studies of FGR Models corroborate with Literature Evidence}

Models incorporating domain-specific chemical knowledge offer the potential for interpretable reasoning, enhancing the framework's predictive robustness and user trust. This work demonstrates that models constructed using the proposed Functional Group Representation (FGR) framework yield interpretable predictions by systematically identifying functional groups that contribute meaningfully to molecular properties. The methodology employed for interpretability analysis is detailed in~\cref{interpret_methodology}.

Based on functional group representations, our interpretability analysis demonstrates that the model captures universal and endpoint-specific structural features critical for accurate molecular property prediction. By analyzing importance rankings across 14 diverse datasets, the model consistently assigns high attribution scores to chemically meaningful substructures such as alcohols, aromatic systems, nitrogen heterocycles, sp$^3$-hybridized carbon atoms, and tertiary amines. These functional groups are known to influence molecular recognition through mechanisms such as hydrogen bonding, $\pi$-$\pi$ interactions~\cite{brylinski_aromatic_2018}, molecular flexibility~\cite{kombo_3d_2013}, and electrostatic effects~\cite{guan_triazoles_2024}, aligning with foundational principles in medicinal chemistry. Beyond these universal patterns, the model uncovers distinct feature preferences tied to specific prediction tasks. In ADMET-related datasets (BACE, BBBP, ClinTox, SIDER, Tox21, and ToxCast), the model assigns high importance to halogenated aromatics~\cite{gentry_effect_1999} and reactive carbonyl groups~\cite{schultz_trends_2004}, consistent with their established roles in metabolic stability and toxicity, respectively. In the bioactivity-focused datasets 746\_aa and \textit{E.~coli}, feature attribution analysis highlights the prominence of peptide-like motifs (MFG patterns) and specific SMARTS-defined substructures resembling peptidomimetic antibiotics, which are known to facilitate membrane disruption and protein target engagement in bacterial systems~\cite{domalaon_short_2016}. Similarly, the cancer cell line datasets (A2780, CCRF-CEM, and DU-145) prioritize heterocyclic scaffolds, such as pyridine rings, which are widely recognized for their roles in kinase binding and enzyme inhibition in oncology~\cite{mohamed_medicinal_2021}. Additionally, these datasets emphasize on peptide-like motifs potentially capturing some patterns which might be essential for anticancer activity. In contrast, physicochemical property prediction tasks (FreeSolv, ESOL and Lipop) prioritize functional groups associated with solubility, lipophilicity, and hydrogen bonding most notably alcohols, ethers, and carboxylic acids~\cite{loeffler_hydration_2019}. In addition to traditional functional groups, molecular descriptors such as \texttt{Ipc} and \texttt{BertzCT}, which capture molecular complexity and topological features, frequently appear among high-attribution features present in 11 datasets. These findings demonstrate that the FGR framework recovers canonical structure-activity relationships and provides biologically meaningful, dataset-specific explanations. The analysis enhances confidence in its application to cheminformatics and drug discovery tasks by improving model transparency and interpretability.

Next, we validate our interpretability findings through supporting evidence from the scientific literature, using representative case studies on the BACE, BBBP, FreeSolv, and \textit{E.~coli} datasets. Additional case studies are provided in the Supplementary Information S7. Moreover, the top 10 functional groups with corresponding attribution scores for each dataset are presented in Supplementary Information S7.1, while a more comprehensive list of the top 50 functional groups is included in S7.2. Functional group frequency distributions across datasets are summarised in S7.3. These analyses underscore the framework's capacity to deliver biologically meaningful and interpretable insights.

\begin{figure}
    \centering
    \includegraphics[width=\textwidth]{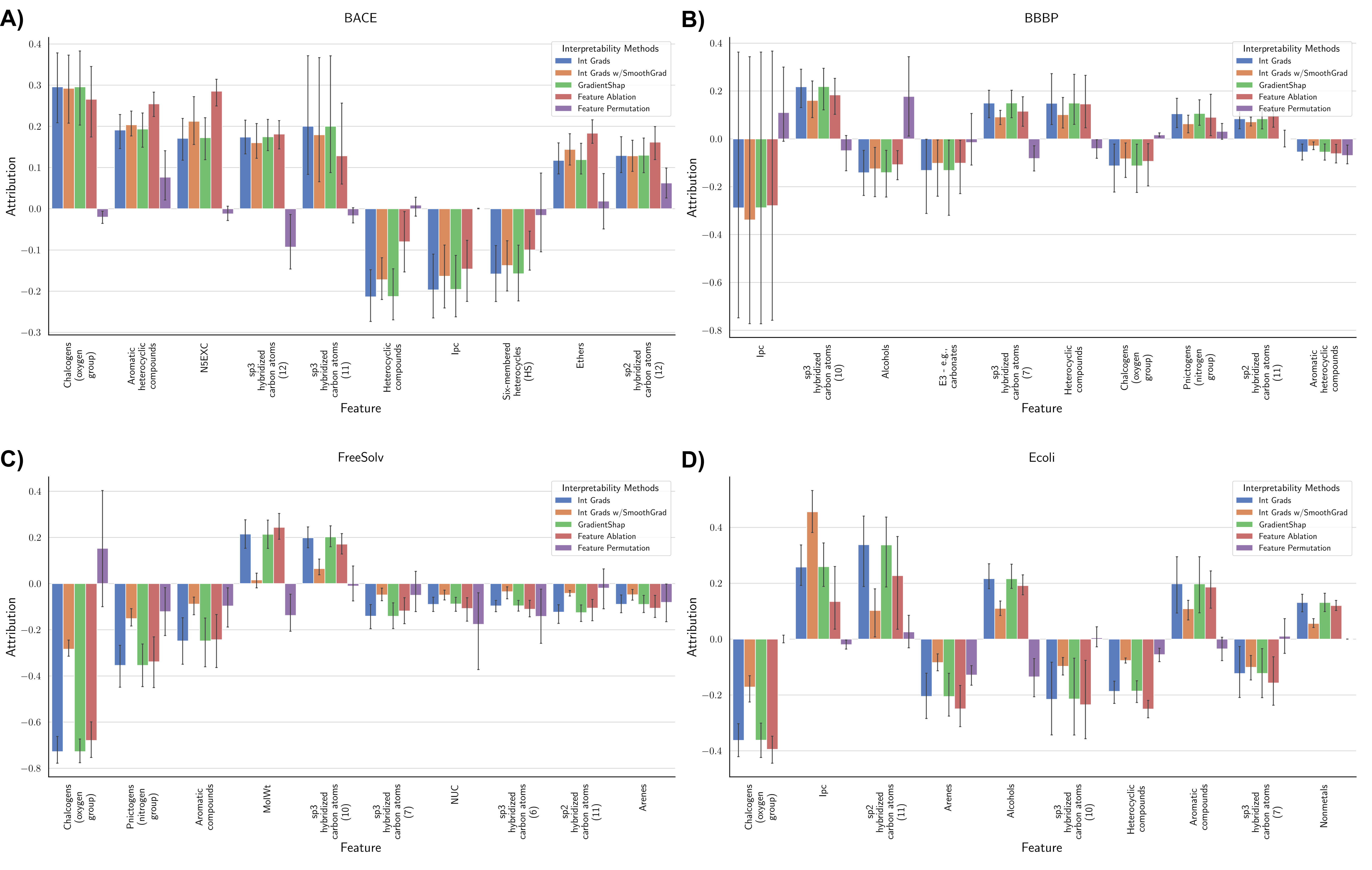}
    \caption{Interpretability analysis on \textbf{A)} BACE, \textbf{B)} BBBP, \textbf{C)} FreeSolv and \textbf{D)} \textit{E.~coli} datasets. The attribution scores were obtained using five different attribution algorithms averaged across five folds.}
    \label{fig:interpret}
\end{figure}

\subsection{Functional Groups affecting \texorpdfstring{$\beta$}{beta}-Secretase 1 Inhibition}
A primary therapeutic strategy for Alzheimer's disease has focused on inhibiting the enzyme $\beta$-Secretase 1 (BACE1), crucial in forming and aggregating amyloid-beta peptides. To this end, chemists have explored a variety of structural chemotypes to develop effective BACE1 inhibitors. The functional groups with the top 10 attribution scores and the top 50 attribution scores are provided in~\cref{fig:interpret}(A). Extensive literature evidence supports the role of chalcogens as prominent contributors to BACE1 inhibition~\cite{narayanan_flavonoid_2024}, consistent with the high positive attribution scores observed in our model (see~\cref{fig:interpret} (A)). The computational framework in the literature also suggests that aromatic heterocycles (the functional group with the top-10 attribution score) may enhance BACE1 inhibition via interactions within the enzyme's active site, as exemplified by aminothiazoline- and amino oxazoline-based inhibitors~\cite{mureddu_fragment-based_2022,marin_new_2022}. 
Furthermore, the presence of sp$^3$-hybridized carbon atoms is associated with positive model attributions, aligning with contemporary drug discovery strategies that emphasise conformational restriction by incorporating sp$^3$-rich scaffolds. Studies on cyclopropane-containing BACE1 inhibitors indicate that rigid sp$^3$ centres can induce alternative binding conformations and enhance inhibitory potency~\cite{yonezawa_conformational_2012}. Building upon prior findings related to the roles of carboxylic acid moieties in norstatine- and tert-hydroxyl group-based inhibitors, the FGR-based model in this work similarly predicts that these functional groups confer advantages in peptidomimetic BACE1 inhibitors. The effect is likely attributable to improved hydrophilic interactions and optimised hydrogen bonding within the enzyme's active site~\cite{ghosh_bace1_2014,kimura_design_2006}. Additionally, the framework underscores the potential contribution of the pyridine ring, representing both tertiary amines and aromatic systems, in mediating BACE1 inhibition. Specifically, 2-aminopyridine-based inhibitors have demonstrated favourable interactions with the S2' subpocket and Trp76 residue~\cite{ghobadian_novel_2018}.

\subsection{Functional Groups affecting Blood-Brain Barrier Penetration}
The drug molecules that can traverse through the blood-brain barrier (BBB) are important for treating central nervous system (CNS) disorders. The general chemical modification strategy to generate viable candidates is to modify the polarity and lipophilicity of the parent drugs. The interpretability analysis of the FGR-based model for the BBBP dataset, as shown in~\cref{fig:interpret}(B) (top 10 functional groups), indicates that pnictogens (nitrogen‐containing functional groups) receive the highest positive attribution scores, in agreement with experimental evidence showing that protonatable nitrogen atoms facilitate organic‐molecule permeation across biological barriers under physiological conditions~\cite{rosa_identifying_2024}. Furthermore, the model (\cref{fig:interpret}(B))  also assigns near-neutral to slightly positive attribution scores to aromatic heterocyclic compounds. The finding is consistent with existing literature, indicating that nitrogen-containing structures and aromatic rings are more frequently observed in  BBB-permeable compounds than non-permeable ones. In contrast, the model exhibits mixed attribution patterns for oxygen-containing functional groups. Alcohols tend to receive slightly negative attributions, while chalcogen-containing groups are generally assigned more positive values. This observation reflects the nuanced role that oxygen-bearing moieties play in BBB permeability. Specifically, hydroxyl groups (–OH) are known to facilitate permeability via hydrogen bonding interactions with BBB components. However, the negative attribution associated with alcohols may reflect the influence of multiple hydroxyl groups, which can increase molecular polarity. Compounds with large polar surface areas are less likely to permeate the BBB, with an estimated upper limit ranging from 60 to 90 Å$^2$~\cite{mikitsh_pathways_2014}. Finally, the model assigns consistently positive attribution to sp$^3$-hybridized carbon atoms, aligning with principles in medicinal chemistry. Empirical studies in pharmaceutical optimisation have shown that both the fraction of sp$^3$-hybridized carbon atoms (Fsp$^3$) and the number of stereocenters tend to increase as compounds are refined for better pharmacokinetic and pharmacodynamic properties~\cite{beckers_25_2022}.

\subsection{Functional Groups affecting Solubility}
Chemical solubility is a fundamental and uncomplicated chemical feature based on well-established first-principles knowledge. The different parts of a chemical compound, such as functional groups, can be divided into two categories: hydrophilic or hydrophobic. Hydrophilic groups, like alcohols, amines, and carboxyls, strongly attract water and can improve the overall solubility of a substance. These groups typically contain atoms other than carbon, such as nitrogen and oxygen. Conversely, hydrophobic groups, which mainly consist of carbon-based chains, rings, and halogens (chlorine, bromine, iodine), tend to decrease the solubility of a chemical and are regarded as `water-repelling'. The interpretability analysis of our FGR model (refer to~\cref{fig:interpret}(C) (top 10 functional groups) reveals consistent and chemically meaningful patterns in attributing molecular features to water solubility. Oxygen-containing functional groups exhibit the most prominent negative attribution scores across all interpretability methods, as seen in~\cref{fig:interpret}(C). This observation aligns with well-established chemical principles, as such groups, particularly hydroxyl functionalities, are known to enhance water solubility through hydrogen bond formation with water molecules. Similarly, nitrogen-containing functional groups also demonstrate negative attribution scores, suggesting a contribution to increased solubility. This result corroborates the documented solubility-enhancing properties of amines and related nitrogen-based functionalities. The molecular weight feature shows positive attribution, correctly capturing the inverse relationship between molecular size and solubility. This trend is consistent with the well-established principle that solubility decreases with increasing hydrocarbon chain length. Features associated with sp$^3$-hybridized carbon atoms consistently display positive attribution scores. This finding reflects the hydrophobic character of aliphatic carbon chains, which diminishes solubility by increasing the non-polar surface area that must be accommodated in aqueous environments. Adding each methylene group further reduces water solubility due to enhanced hydrophobic interactions. Lastly, the acetate functional group shows negative attribution scores, aligning with carboxylate-containing moieties' known hydrophilic nature. Carboxylic acids and their conjugate bases are widely recognised for enhancing aqueous solubility via ionic interactions and hydrogen bonding~\cite{loeffler_hydration_2019}.

\subsection{Functional Groups affecting Antibiotic Activity}
A comprehensive understanding of the physicochemical properties inherent to the antibiotic chemical space is essential to address the challenge of antibiotic resistance. Such knowledge is vital for informing and guiding the development of new antibiotic agents, facilitating the identification of promising candidates with enhanced efficacy and resistance profiles. By leveraging the link between the functional groups and these properties, researchers can optimise antibiotic design strategies and foster the discovery of urgently needed antimicrobial therapies.~\cref{fig:interpret}(D) provide the top 10 and top 50 functional groups identified by the interpretability analysis of the FGR models for the {\it E. coli} datasets. The interpretability analysis of our model on the \textit{E. coli} dataset as shown in reveals several key molecular features associated with antibacterial activity, many of which are well-supported by existing experimental evidence~\cite{sen_escherichia_2023,mhussein_synthesis_2022,zhang_natural_2021,islam_recent_2023,kock_1-hydroxy-1-norresistomycin_2005,pawlowski_evolving_2016}. Alcohols exhibit positive attribution scores in our model, consistent with experimental findings that demonstrate a chain-length dependent toxicity of alcohols against \textit{E. coli}. Specifically, alcohol toxicity increases exponentially with chain lengths ranging from 2 to 6 carbon atoms~\cite{sen_escherichia_2023}. This observation supports the model's attribution patterns and highlights the relevance of alcohol chain length in modulating bacterial inhibition. Heterocyclic compounds also show strong positive attribution scores (see~\cref{fig:interpret}(D)), aligning with extensive pharmacological studies. Nitrogen-containing heterocycles exhibit bioactivity against various pathogens, and metal complexes derived from these scaffolds have been explored for their broad pharmacological potential~\cite{mhussein_synthesis_2022}. The model's identification of these compounds as important contributors to antibacterial activity underscores the utility of heterocycles in antimicrobial drug design. The model further attributes high positive scores to aromatic compounds, reflecting their established role in antibacterial mechanisms. Phenolic compounds, characterised by aromatic and hydroxyl functionalities, are well-documented for their antimicrobial effects. For instance, compound phenolic acid (CPA) 19 demonstrated superior efficacy against \textit{E. coli} compared to other phenolic combinations~\cite{zhang_natural_2021}. Pyridine scaffolds are recognised for their structural versatility and ability to modulate interactions with biological targets. Numerous pyridine-containing drugs are FDA-approved and listed in major pharmaceutical databases. A notable example is sulfapyridine, an antibacterial agent synthesised by linking pyridine to sulfanilamide, which has shown substantial efficacy in treating bacterial infections~\cite{islam_recent_2023}. Aliphatic ethers also emerge as positively contributing features, likely due to their ability to enhance bioavailability and improve membrane permeability of antibiotic molecules~\cite{kock_1-hydroxy-1-norresistomycin_2005}. The model strongly emphasises the presence of $\beta$-lactam moieties—a hallmark of many clinically significant antibiotics. These structures inhibit bacterial cell wall synthesis, making them essential in treating microbial infections~\cite{pawlowski_evolving_2016}. The analysis indicates that the FGR model captures essential functional groups for antibiotic activity. 

\section{Methodology}\label{methodology}

\subsection{Problem Settings}\label{problem_setting}
The molecular property prediction problem involves mapping each molecule to a set of properties of size $k$ (depending on the number of tasks) and treating it as a classification ($y \in \{0,1\}^k$) or regression ($y \in \mathbb{R}^k$) problem. Molecule representation learning is an important task in developing models for the property prediction task. In this work, we aim to learn a latent embedding vector $\zbf \in \mathbb{R}^l$ ($l$ is a hyperparameter) for each molecule from the available chemical structures and descriptors and use it in different downstream property prediction tasks. The SMILES (Simplified Molecular Input Line Entry System) line notation represents a chemical structure in a way that the computer can use. We use a set of SMILES strings for $n$ molecules, $\mathcal{S}=\{S_i \mid i \in n\}$, where each $S_i$ is associated with a representation $\zbf_i$, learnt using an encoder function $f_{\mathrm{e}}: \mathcal{S}\rightarrow\mathbb{R}^l$ based on a feedforward neural network.

As input to the encoder, a functional group vocabulary using the string set ($\mathcal{S}$) is curated using the ToxAlerts~\cite{sushko_toxalerts_2012} web server and molecules in the PubChem~\cite{kim_pubchem_2016} database. The latent embedding vector ($\zbf_{\mathrm{G}}$) is learnt using the functional group representation and an autoencoder~\cite{hinton_reducing_2006}. Further, we also consider 2D molecular descriptors ($\zbf_{\mathrm{DE}}$) calculated using RDKit~\cite{landrum_rdkit_2013} along with the learned latent embedding ($\zbf_{\mathrm{G}} \oplus \zbf_{DE}$) for understanding its role in the property prediction task and improving downstream performance. For details on the model architecture, refer to Supplementary Information S8; for hyperparameter tuning, see Supplementary Information S9; and for the number of learnable parameters, see Supplementary Information S10.

\subsection{Generation of Functional Group Vocabulary}\label{fg_vocab}
This section explains the generation of functional group vocabulary inspired by chemistry. A molecule in chemistry comprises substructures that impart distinct chemical, physical, and biological properties. The substructures are labelled as functional groups, consisting of a few atoms, typically carbon and hydrogen, along with one or more heteroatoms such as oxygen, nitrogen, sulfur, or halogens (like chlorine or bromine). Functional groups can also be termed reaction centres, and different functional groups are associated with different sets of properties like melting point, solubility and nucleophilicity. We construct a comprehensive vocabulary of functional groups through two approaches: (i) curating functional groups identified and cataloged by chemists (denoted as FG) from the Toxalerts web server, and (ii) applying a sequential pattern mining algorithm to a large molecular corpus to identify and extract Mined Functional Groups (MFG). This dual approach, as illustrated in Fig~\ref{fig:introduction_methodology}, allows us to combine both established and newly discovered functional groups, offering a broader and more nuanced representation of molecular structures.

\subsubsection{Functional Groups Curated from ToxAlerts}
In this study, we use the ToxAlerts web server, which collects and stores toxicological structural alerts from literature defined and verified by chemists in the SMARTS~\cite{noauthor_daylight_nodate} format (an extension of the SMILES representation). The substructures are based on patterns and are much easier to interpret, as each substructure is associated with a mechanism of action for different toxicological endpoints. Let $\mathcal{FG}=\{ \mathrm{FG}_1,\ldots, \mathrm{FG}_a\}$ denote a set of functional groups curated from the web server. We only take into account verified alerts and valid SMARTS strings. The final vocabulary contains 2672 functional groups and any molecule $S_i\in\mathcal{S}$ can be represented by a multi-one-hot encoded vector, $\xbf_{\mathrm{FG}} = [\xbf^{(1)} \,\xbf^{(2)}\ldots\,\xbf^{(a)}]$ where $\xbf^{(i)} = 1$ if $\mathrm{FG}_i \in S$ and $\xbf^{(i)} = 0$, if $\mathrm{FG}_i \notin S$.

\subsubsection{Mined Functional Groups from PubChem}
Let $S_i\in \mathcal{S}$ be the SMILES string of an $i$th molecule (or molecular graph) in the PubChem database, and $C$ be a consecutive sub-string of $S_i$. Then, $C$ corresponds to a depth-first traversal of a molecular sub-graph. $C$ is a frequent substructure or mined functional group if its occurring frequency is above a threshold \(\eta\). The method assumes that the same SMILES sub-strings will represent sub-structures that appear across different molecules, and hence, it is possible to mine frequent substructures through a SMILES sub-string-based approach. We look for frequent patterns in SMILES of molecules ($>$114 Million) available in the PubChem database using a Chemical Sequential Pattern Mining (SPM)~\cite{huang_caster_2020} algorithm with an appropriate frequency threshold \(\eta\) and maximum vocabulary size ($\mathrm{MVS}$).

Let \(\mathcal{MFG} = \{\mathrm{MFG}_1, \cdots, \mathrm{MFG}_b\}\) denote the set of frequent sub-structures identified by applying the sequential pattern mining algorithm. Any molecule $S\in\mathcal{S}$ can be represented by a multi-one-hot encoded vector, $\xbf_{\mathrm{MFG}} = [\xbf^{(1)} \,\xbf^{(2)}\ldots\,\xbf^{(b)}]$ where $\xbf^{(i)} = 1$ if $\mathrm{MFG}_i \in S$ and $\xbf^{(i)} = 0$, if $\mathrm{MFG}_i \notin S$. In this work, we set $\eta=500$ and $\mathrm{MVS}=30000$ to ensure the common SMILES substrings can be included in the vocabulary. Lowering $\eta$ increases the number of identified patterns, causing the vocabulary size to reach its maximum limit.

\begin{algorithm}
\caption{Sequential Pattern Mining Algorithm}\label{algo1}
\begin{algorithmic}[1]
\Require $\mathrm{MVS}, \eta > 0$ 
\State Initialize $\mathcal{MFG}$ to set of atoms and bonds and $\mathbb{V}$ is the set of tokenized SMILES strings with corresponding frequencies
\For{$t = 1 \ldots b$}
    \State $\textsc{(A, B), freq} \leftarrow \mathrm{scan}\ \mathbb{V}$
     \If{$\textsc{freq} < \eta$}
        \State $\mathrm{break}$ 
    \Else{}
        \State $\mathbb{V} \leftarrow \mathrm{find} \textsc{(A, B)} \in \mathbb{V},  \mathrm{replace}~\mathrm{with}\  \textsc{(AB)}$
        \State $\mathcal{MFG} \leftarrow \mathcal{MFG} \cup \textsc{(AB)}$
    \EndIf
\EndFor
\end{algorithmic}
\end{algorithm}

\subsection{Latent Feature Embedding in FGR Framework}
\label{latent_feature}
In the initial step, we obtain $\xbf_{\mathrm{FG}}$ and $\xbf_{\mathrm{MFG}}$ for each molecule using the vocabularies $\mathcal{FG}$ and $\mathcal{MFG}$, respectively. In the second step, we obtain a lower-dimensional latent feature encoding using an autoencoder to generalise the FGR framework-based representations to new molecules and the downstream property prediction tasks. The framework uses different input representations based on the vocabulary: (i) FG representation ($\xbf_{\mathrm{FG}}$), (ii) MFG representation ($\xbf_{\mathrm{MFG}}$) and (iii) Combined Representation ($\xbf_{\mathrm{FG}}\oplus \xbf_{\mathrm{MFG}}$). The objective here is to learn functions $f_{\xbf_{\mathrm{G}}}: \xbf_{\mathrm{G}} \rightarrow \mathbb{R}^l$ using autoencoders where $\xbf_{\mathrm{G}}$ is a multi-hot vector of appropriate dimension (say $p$) depending on the input representation. The main advantage of $f_{\xbf_{\mathrm{G}}}$ is that it can be decoupled from the downstream prediction tasks and learned in an unsupervised manner with unlabeled data. Optionally, the 2D descriptors can also be concatenated with $\zbf_{\mathrm{G}}$ for further property prediction tasks. Including 2D descriptors is a hyperparameter dependent on the property prediction tasks. In this work, we employ autoencoders to handle $f_{\xbf_{\mathrm{G}}}$, and we will now proceed to describe the components of the autoencoders, including the encoder and decoder and the reconstruction loss function.

\begin{itemize}
    \item  \textbf{Encoder}: A neural network (NN) is applied to each of the functional group representations \(\xbf_{\mathrm{G}} \in \{0,1\}^p,\) of molecules. Using weight \(\Wbf_{\mathrm{e}}\) and bias \(\bbf_{\mathrm{e}}\), then, the encoder can be expressed as:
\begin{equation}
    \zbf_{\mathrm{G}} = \Wbf_{\mathrm{e}}\xbf_{\mathrm{G}} + \bbf_{\mathrm{e}}
\end{equation}
where $\zbf_{\mathrm{G}} \in \mathbb{R}^l$ is a latent feature vector. 
\item \textbf{Decoder}: To measure the information retention of the latent representation, $\zbf_{\mathrm{G}}$, the reconstruction of the input \(\xbf_{\mathrm{G}}\) using the decoder using an another NN with weight \(\Wbf_{\mathrm{d}}\) and bias \(\bbf_{\mathrm{d}}\) is performed as follows:
\begin{equation}
\hat{\xbf}_{\mathrm{G}} = \sigma(\Wbf_{\mathrm{d}}\zbf_{\mathrm{G}} + \bbf_{\mathrm{d}})  
\end{equation}
where the $\sigma(\cdot)$ is the element-wise sigmoid function defined as \(\sigma(a) = 1/(1+e^{-a})\).
\item \textbf{Tied-weight Autoencoder}: The weights of the autoencoder can optionally be tied to make the autoencoder well-posed ($\Wbf_{\mathrm{d}} = \Wbf_{\mathrm{e}}^\top$). Tied weight autoencoders are easier to train with fewer parameters to learn and act as a form of regularisation. 
\item \textbf{Uncorrelated Bottleneck Constraint}: Penalising the sum of off-diagonal elements of the encoded features covariance can make the autoencoder well posed, making it easier to optimise. Uncorrelated feature encoding can be achieved by minimising the following loss function:
\begin{equation}
    L_{ubc}(\zbf_{\mathrm{G}}) = \sum^{p \times p}_{i=1}(\Cov(\zbf_{\mathrm{G}}) - \diag(\Cov(\zbf_{\mathrm{G}})))^2
\end{equation}
\item \textbf{Reconstruction Loss Function}:
The weights and biases of the encoder-decoder, \(\Wbf_{\mathrm{e}}\), \(\Wbf_{\mathrm{d}}\), \(\bbf_{\mathrm{e}}\) and  \(\bbf_{\mathrm{d}}\), are learnt by minimizing the reconstruction loss ($L_r$) between $\xbf_{\mathrm{G}}$ and $\hat{\xbf}_{\mathrm{G}}$ as follows:
\begin{align}
\mathrm{BCE}(\xbf_{\mathrm{G}},\hat{\xbf}_{\mathrm{G}}) &= \sum_{i=1}^p(\xbf_{\mathrm{G}}^{(i)}\log(\hat{\xbf}_{\mathrm{G}}^{(i)}) + (1 - \xbf_{\mathrm{G}}^{(i)})\log(1 - \hat{\xbf}_{\mathrm{G}}^{(i)}))   \\
p_t &= \exp(-\mathrm{BCE}(\xbf_{\mathrm{G}},\hat{\xbf}_{\mathrm{G}})) \\
L_r(\xbf_{\mathrm{G}},\hat{\xbf}_{\mathrm{G}}) &= -\alpha_t(1-p_t)^\gamma\log(p_t)
\label{Eq:ReconLoss} 
\end{align}
where $\xbf_{\mathrm{G}}^{(i)}$ denotes the $i$th element of $\xbf_{\mathrm{G}}$.  
\end{itemize}
We use the Focal Loss~\cite{lin_focal_2017} typically used in dense object detection tasks to handle the high-class imbalance (vector sparsity) present in the feature representation. $\alpha_t$ balances the importance of positive/negative examples, while $\gamma$ helps differentiate between easy/hard to classify examples. $\alpha_t$ and $\gamma$ are hyperparameters set using cross-validation. 

Depending on $\xbf_{\mathrm{G}}$, we develop three types of feature representation as described in Figs
\begin{itemize}
    \item \textbf{Functional Group (FG) Representation}: In this representation, each molecule is represented by the functional groups curated from the ToxAlerts web server. Here, a molecule is converted to a multi-hot encoding vector, $\xbf_{\mathrm{G}}=\xbf_{\mathrm{FG}} \in \{0,1\}^a$ with $p=a$, and the corresponding latent embedding (or feature) vector, $\zbf_{\mathrm{G}}=\zbf_{\mathrm{FG}}$ that is obtained by applying an autoencoder as shown in~\cref{fig:introduction_methodology}. 
    \item \textbf{Mined Functional Group (MFG) Representation}: Each molecule is first represented by a set of mined functional groups obtained by applying the SPM algorithm to the PubChem database. A molecule is represented by a multi-hot encoding vector, $\xbf_{\mathrm{G}}=\xbf_{\mathrm{MFG}}\in \{0,1\}^b$ with $p=b$, and the corresponding latent feature vector, $\zbf_{\mathrm{G}}=\zbf_{\mathrm{MFG}}$ that is obtained by applying an  autoencoder as shown in~\cref{fig:introduction_methodology}.
    \item \textbf{Combined Representation}: This approach uses functional groups curated from the ToxAlerts web server and the mined functional groups from the PubChem database to learn the latent embedding. A molecule is represented by concatenation of multi-hot encoding vectors by the FG and MFG representations, i.e., $\xbf_{\mathrm{G}}=\xbf_{\mathrm{FG}} \oplus \xbf_{\mathrm{MFG}} \in \{0,1\}^{a+b}$ with $p=a+b$. The corresponding latent feature vector is defined as $\zbf_{\mathrm{G}}=\zbf_{FGR}$ that is obtained by applying an autoencoder on  $\xbf_{\mathrm{FG}} \oplus \xbf_{\mathrm{MFG}}$  as shown in Fig~\ref{fig:introduction_methodology}.
    \item \textbf{RDKit Descriptors}: The RDKit library calculates 2D descriptors such as molecular weight, charge and number of electrons for each molecule. The descriptors are of different scales, so $L_2$ normalisation is done over the feature dimension for stable pipeline training. The descriptors calculated ($\zbf_{\mathrm{DE}}$) are of size 211, and the final latent embedding is generated by concatenating any of the above representations with the descriptors. The full list of descriptors used for calculation is provided in Supplementary Information S11.
\end{itemize}

\subsection{Property Prediction Task}
\label{prediction_task}
 In the previous step, molecular functional group representations \(\xbf_{\mathrm{G}} \in [0,1]^p\) and its corresponding latent feature encoding \(\zbf_{\mathrm{G}} \in \mathbb{R}^l\) are obtained for different types of functional group representations. As shown in~\cref{fig:introduction_methodology}, the next step is to use the latent feature encoding for predicting the properties of molecules. The property prediction is performed by building an appropriate model between the latent feature vector \(\zbf_{\mathrm{G}}\) and the property of the interest. Here, a fully connected neural network with the weight matrix \(\Wbf_f\) and bias vector \(\bbf_f\) is used to predict the property ($\hat{y}$) based on $\zbf_{\mathrm{G}}$. The prediction step is defined as:
\[\hat{y} = \mathrm{act}(\Wbf_f\zbf_{\mathrm{G}} + \bbf_f)\] 
$\mathrm{act} = \sigma$ if classification and no activation for regression.
The weights \(\Wbf_f\) and biases \(\bbf_f\) are optimised by minimising the binary cross-entropy loss for the classification case and the smooth $L_1$ loss for the regression case. 

The total loss $L_t$  is minimised during the training phase as follows:
\begin{equation}
    L_{t}= \sum\, (L_{\mathrm{e}}(\xbf_{\mathrm{G}},y) +\alpha L_r(\xbf_{\mathrm{G}},\hat{\xbf}_{\mathrm{G}}) + \beta L_{ubc}(\zbf_{\mathrm{G}}))
\end{equation}
where $\alpha$ and $\beta$ are hyperparameters.
The total loss $L_t$ can be minimised by assigning weights to the loss terms according to the prediction task. The model can be trained end-to-end with labelled molecules alone or combined with unlabeled data to conduct unsupervised pre-training.

\subsection{Experimental Setup}
\label{experimental_setup}
\subsubsection{Dataset Splitting}
We evaluate all the models using five independent runs across different seeds and report the average results. We split each dataset into training, validation, and testing sets with a ratio of 0.8/0.1/0.1 using random and scaffold splits. Scaffold splitting results in structurally different splits to better estimate the model's performance. Splitting data based on these scaffolds~\cite{bemis_properties_1996} ensures molecules with the same core structure never appear in both the training and test sets, forcing the model to learn generalizable patterns applicable to unseen scaffolds. This is crucial for tasks like predicting the activity or properties of novel molecules outside the training data.

\subsubsection{Baselines}\label{baselines}
We evaluated the suggested framework against several state-of-the-art baseline models on benchmark datasets for predicting molecular properties. The baseline models included GNNs with and without pre-training, sequence-based models, models that utilize 3D geometry information, and knowledge graphs. The following models were used as baselines:
\begin{itemize}
    \item GCN~\cite{kipf_semi-supervised_2017}: Graph Convolutional Networks (GCNs) leverage graph structures of molecules to encode atom interactions, capturing crucial spatial and bonding information for accurate property prediction.
    \item MPNN~\cite{gilmer_neural_2017}: Message Passing Neural Networks (MPNNs) iteratively exchange information between atoms, mimicking real-world chemical interactions for rich property prediction. 
    \item GIN~\cite{xu_how_2018}: Graph Isomorphism Network (GIN) is a permutation-invariant representation that excels at handling diverse structures and identifying similar molecules, even with different atom arrangements.
    \item N-GRAM~\cite{liu_n-gram_2019}: Primarily used in natural language processing, N-Gram is a pre-trained model that captures snippets of text (1-3 words) to understand sequences and patterns. 
    \item DMPNN~\cite{yang_analyzing_2019}: Directed Message Passing Neural Networks (DMPNNs) use message flow along bonds, allowing the model to focus on the specific nature of each bond and its influence on properties.
    \item CMPNN~\cite{song_communicative_2020}: The message interactions between nodes and edges are strengthened through a communicative kernel in the Communicative Message Passing Neural Network (CMPNN) to enhance molecular embedding. 
    \item GROVER~\cite{rong_self-supervised_2020}: GROVER uses Message Passing Networks and Transformer-style architecture to create more expressive molecule encoders, incorporating two self-supervised tasks.
    \item MGSSL~\cite{zhang_motif-based_2024}: Motif-based Graph Self-supervised Learning (MGSSL) introduces a novel self-supervised motif generation framework in which GNNs are asked to make topological and label predictions.
    \item GEM~\cite{fang_geometry-enhanced_2022}: Geometry Enhanced Molecular Representation (GEM) is a framework designed to learn the geometry of molecules based on a self-supervised approach at the geometry level.
    \item GraphMVP~\cite{liu_pre-training_2021}: The Graph Multi-View Pre-training (GraphMVP) framework uses self-supervised learning (SSL) to learn from 2D topological structures and 3D geometric views.  
    \item MolCLR~\cite{wang_molecular_2022}: Molecular Contrastive Learning of Representations via Graph Neural Networks (MolCLR) is a self-supervised learning framework that uses graph neural networks and large unlabeled data to predict molecular properties.
    \item KANO~\cite{fang_knowledge_2023}: Knowledge graph-enhanced molecular contrastive learning with functional prompt (KANO) exploits an element-oriented knowledge graph as a prior in pre-training and learns functional prompts in fine-tuning for downstream property prediction tasks. 
\end{itemize}

\subsubsection{Model Training}
\label{model_training}
Since each descriptor's scale and distribution might differ, RDKit descriptors are normalized to a $[0,1]$ range using the $L_2$ normalization. We use the Stochastic Gradient Descent~\cite{sutskever_importance_2013} (SGD) optimizer along with Sharpness Aware Minimization~\cite{foret_sharpness-aware_2020} (SAM) to train the model for better generalization with a batch size of 16. All the experiments were carried out using four A100 GPUs with bf16 mixed precision for 50 training epochs implemented in PyTorch.

\subsubsection{Performance Evaluation}
\label{performance_evaluation}
For MoleculeNet datasets, as suggested we use the macro averaged receiver-operating characteristic-area-under-the-curve~\cite{bradley_use_1997} (ROC-AUC) metric for evaluating the binary classification tasks (BBBP~\cite{martins_bayesian_2012}, Tox21~\cite{noauthor_tox21_nodate}, ToxCast~\cite{richard_toxcast_2016}, SIDER~\cite{kuhn_sider_2016}, ClinTox~\cite{gayvert_data-driven_2016}, BACE~\cite{subramanian_computational_2016}, MUV~\cite{rohrer_maximum_2009} and HIV~\cite{noauthor_aids_nodate}). For regression tasks, we use the root mean squared error (RMSE) for ESOL~\cite{delaney_esol_2004}, FreeSolv~\cite{mobley_freesolv_2014} and Lipophilicity~\cite{gaulton_chembl_2012} tasks and the mean absolute error (MAE) for quantum mechanics datasets (qm7~\cite{blum_970_2009}, qm8~\cite{ramakrishnan_electronic_2015}, qm9~\cite{ramakrishnan_quantum_2014}). We use the $\mathrm{R}^2$ metric (higher scores are better) to evaluate the KekuleScope~\cite{cortes-ciriano_kekulescope_2019} and LMC~\cite{wenzel_predictive_2019} regression datasets, and the RMSE for Malaria~\cite{xiong_pushing_2020} dataset. We use the ROC-AUC metric for evaluation performance for CYP~\cite{li_prediction_2018} and peptide cleavage datasets (\textit{E.~coli}~\cite{stokes_deep_2020},  Mpro~\cite{douangamath_crystallographic_2020}, Schilling, Impens, 1624\_aa, 746\_aa~\cite{rognvaldsson_state_2015}).

\subsection{Interpretability Analysis}\label{interpret_methodology}
Interpretability studies were carried out using the Captum~\cite{kokhlikyan_captum_2020} library for Pytorch. The library offers different attribution algorithms and three types of attribution variants: primary attribution, neuron attribution, and layer attribution. We use primary attribution methods like Integrated Gradients~\cite{sundararajan_axiomatic_2017}, GradientShap~\cite{lundberg_unified_2017}, Feature Permutation~\cite{fisher_all_2019} and Feature Ablation~\cite{zeiler_visualizing_2014} to obtain the feature-level importance of functional groups and descriptors. A crucial aspect of attribution analysis is the choice of baseline, which serves as a reference input against which the contributions of features are measured. The baseline is typically chosen to represent the absence of meaningful input information. In our study, we define the baseline as an input vector of zeros, corresponding to the absence of all functional groups. Attribution scores are calculated for each feature based on its contribution to the model's predictions. For methods like Integrated Gradients, this involves accumulating gradients along a path from the baseline to the actual input. For other methods, such as Feature Ablation, features are systematically removed or replaced with baseline values to assess their individual impact.

To ensure robust and reliable attribution results, we average the scores obtained from models trained on multiple cross-validation folds. This averaging mitigates the variability introduced by model initialization and training, providing a more stable estimate of feature importance. Finally, the computed attribution scores are visualized using grouped bar plots, offering insights into the relationship between molecular substructures and their associated properties.

\section{Conclusions}
This study presents a functional group representation (FGR) framework using the concept of functional groups in chemistry for molecular representation learning. The proposed FGR framework-based molecular embeddings have been evaluated on several benchmark datasets. The framework performs at par and sometimes better than the state-of-the-art algorithms in classification and regression tasks. The model's representations align well with the established chemical understanding of functional group behaviour. The alignment analysis based on scaffolds (clustering of molecules based on functional groups) on different datasets demonstrated the capture of relevant information. The framework's focus on functional groups enables insights into the rationale behind model predictions, as demonstrated using the BACE and ESOL datasets. Novel insights (new functional groups) were obtained into chemical relationships and properties, which could be explored further to design new molecules. 

Although the framework achieves competitive performance, the representation has some limitations. When used together, the FG and MFG representations have overlapping substructures (bit clash), which might not be desirable. The representation cannot differentiate between structural isomers, a vital defect of the SMILES representation. Future work can explore the effect of pre-training the autoencoder on a large dataset of unlabeled molecules and integrating representations that capture 3D information in the encoding.  

In conclusion, our framework offers a promising approach to molecular representation learning, achieving competitive performance while enhancing interpretability through its grounding in chemical principles. Interpretability studies using functional groups demonstrate the framework's ability to capture meaningful chemical relationships within the learned representations.

\section{Code and Data Availability}

All the scripts to reproduce the results, the datasets used in this work and Supplementary Information are available at \url{https://github.com/bisect-group/fgmolprop}.

\section{Acknowledgements}

Roshan Balaji acknowledges support from the Prime Minister's Research Fellowship (PMRF), India, and Nirav Bhatt acknowledges support from the Ministry of Education, Government of India. Both Roshan Balaji and Nirav Bhatt are also supported by the Wadhwani School of Data Science and AI, IIT Madras. The authors gratefully acknowledge the computational resources provided by Wadhwani School of Data Science and AI and IBSE.

\section{Conflict of Interest}

There are no conflicts of interest to declare.

\bibliographystyle{nature_mi/sn-nature}%
\bibliography{sn-bibliography}%

\end{document}